\documentclass[pdflatex,sn-mathphys-num]{sn-jnl}

\usepackage{graphicx}%
\usepackage{multirow}%
\usepackage{amsmath,amssymb,amsfonts}%
\usepackage{amsthm}%
\usepackage{mathrsfs}%
\usepackage[title]{appendix}%
\usepackage{xcolor}%
\usepackage{textcomp}%
\usepackage{manyfoot}%
\usepackage{booktabs}%
\usepackage{algorithm}%
\usepackage{algorithmicx}%
\usepackage{algpseudocode}%
\usepackage{listings}%

\theoremstyle{thmstyleone}%
\theoremstyle{thmstyletwo}%

\theoremstyle{thmstylethree}%

\begin{document}

\title[Article Title]{Explainable AI-Driven Detection of Human Monkeypox Using Deep Learning and Vision Transformers: A Comprehensive Analysis}


\author[1]{\fnm{Md. Zahid} \sur{Hossain}}\email{zahidd16@gmail.com}
\equalcont{These authors contributed equally to this work.}

\author*[1]{\fnm{Md. Rakibul} \sur{Islam}}\email{rakib.aust41@gmail.com}
\equalcont{These authors contributed equally to this work.}

\author[2]{\fnm{Most. Sharmin Sultana} \sur{Samu}}\email{sharminsamu130@gmail.com}
\equalcont{These authors contributed equally to this work.}

\affil[1]{\orgdiv{Department of Computer Science and Engineering}, \orgname{Ahsanullah University of Science and Technology}, \orgaddress{\city{Dhaka}, \postcode{1208}, \country{Bangladesh}}}

\affil[2]{\orgdiv{Department of Civil Engineering}, \orgname{Ahsanullah University of Science and Technology}, \orgaddress{\city{Dhaka}, \postcode{1208}, \country{Bangladesh}}}


\abstract{Since mpox can spread from person to person, it is a zoonotic viral illness that poses a significant public health concern. It is difficult to make an early clinical diagnosis because of how closely its symptoms match those of measles and chickenpox. Medical imaging combined with deep learning (DL) techniques has shown promise in improving disease detection by analyzing affected skin areas. Our study explore the feasibility to train deep learning and vision transformer-based models from scratch with publicly available skin lesion image dataset. Our experimental results show dataset limitation as a major drawback to build better classifier models trained from scratch. We used transfer learning with the help of pre-trained models to get a better classifier. The MobileNet-v2 outperformed other state of the art pre-trained models with 93.15\% accuracy and 93.09\% weighted average F1 score. ViT B16 and ResNet-50 also achieved satisfactory performance compared to already available studies with accuracy 92.12\% and 86.21\% respectively. To further validate the performance of the models, we applied explainable AI techniques.}




\maketitle

\section{Introduction}\label{sec1}
The monkeypox virus is the source of this uncommon illness \cite{WHO}. It is a member of the same virus family as the Variola virus, which is linked to an increased chance of contracting smallpox. It spreads from animals to humans and can also transfer between individuals and the environment due to its zoonotic nature. After the COVID-19 pandemic, the monkeypox outbreak gained global attention. Frequent symptoms of monkeypox include fever, headache, back pain, muscular aches, exhaustion and enlarged lymph nodes. A rash that can develop on the face, hands, feet, groin and vaginal or anal areas frequently follows these symptoms \cite{WHO}. The presence of inflammation and rash is the primary distinction between monkeypox and related illnesses like chickenpox and measles. These rashes may initially appear mild but can become increasingly severe, itchy and painful over time. However, this distinction is difficult to detect without a Polymerase Chain Reaction (PCR) test.

Medical professionals encounter significant difficulties in rapidly and accurately diagnosing monkeypox. The disease shares symptoms with other viral infections which makes clinical identification challenging. Additionally, access to PCR testing is often restricted particularly in low-resource settings. Despite these challenges, early diagnosis is essential to prevent further transmission and control outbreaks. Monkeypox is primarily transmitted by direct contact with infected animals or humans and by sharing contaminated bedding or clothing. \cite{WHO}. Once exposed, the virus has an incubation period of approximately one to three weeks before symptoms begin to manifest. People with weakened immune systems are particularly vulnerable and have a higher likelihood of severe illness. To reduce transmission, it is critical to follow World Health Organization (WHO) guidelines which emphasize isolation, maintaining physical distance and practicing proper hygiene. While PCR testing remains the gold standard \cite{strick2006diagnostics} for diagnosis, alternative approaches are necessary in regions where such resources are limited. In these settings, image-based analysis and deep learning techniques can serve as effective tools for early detection, aiding in outbreak control and reducing the spread of the virus.

According to recent research, deep learning methods have outperformed and become more versatile than conventional machine learning (ML) models like Random Forest and KNN \cite{sarker2021deep}. Traditional machine learning models rely heavily on domain expertise and human intervention to fine-tune their parameters and improve classification accuracy. Moreover, these models are often constrained to specific tasks for which they were originally developed which limits their applicability in complex and dynamic fields like medical image analysis. Deep learning models, particularly CNNs, have revolutionized various aspects of medical science by enabling automated feature extraction and hierarchical learning \cite{ravi2016deep} \cite{wang2019deep}. Unlike traditional methods, CNNs process images through multiple layers by capturing intricate patterns and learning robust representations \cite{khan2020survey} essential for accurate classification. When trained on large-scale datasets, these models significantly enhance diagnostic accuracy and reliability. 

However, the availability of large amounts of labeled data is necessary for deep learning models to function well, \cite{najafabadi2015deep} which is often scarce in medical research. Transfer learning \cite{weiss2016survey} has become a potent strategy to deal with this issue. Transfer learning makes it possible to effectively adapt to new medical imaging tasks with limited data by utilizing pre-trained models that have already acquired generic image properties from sizable datasets. This approach not only reduces computational costs but also accelerates the training process.

Transfer learning is an effective technique commonly used when there is a shortage of labeled data \cite{weiss2016survey}. This method leverages a pre-trained CNN, such as a model trained on large-scale datasets like ImageNet \cite{deng2009imagenet}, to apply the features it has learned to a more condensed, domain-specific dataset. By reusing the pre-trained model’s feature extraction capabilities, transfer learning improves accuracy, reduces computational requirements and accelerates training compared to building models from scratch.

Monkeypox skin lesions look similar to those of diseases like chickenpox and measles \cite{ali2022monkeypox}. This makes it difficult to classify them accurately. The subtle differences in visual characteristics are difficult to discern using traditional diagnostic methods. Due to this complexity, deep learning models were implemented to improve diagnostic accuracy. These models analyze images at multiple levels and extract important features for accurate disease identification. This helps with early detection and clinical decision-making.

This study utilized the MSLD \cite{ali2022monkeypox} and the MSID \cite{bala2023monkeynet} dataset. Using transfer learning, we put pre-trained deep learning models into practice. The investigated architectures included ResNet-50 \cite{koonce2021resnet}, MobileNet-V2 \cite{sandler2018mobilenetv2}, ViT B16 \cite{dosovitskiy2020image}, CNN-ViT Hybrid, ResNet-ViT Hybrid and ViT from scratch. The models achieved prediction accuracies of 86.21\% for ResNet-50, 93.15\% for MobileNet-V2, 92.12\% for ViT B16, 69.46\% for CNN-ViT Hybrid, 67\% for ResNet-ViT Hybrid and 65.02\% for ViT from scratch.

We integrate eXplainable AI (XAI) techniques to evaluate the system’s capability in detecting diseases from skin lesion images. These methods make decision-making more transparent. They improve the interpretability of the predictions made by deep learning models. Research shows that XAI makes deep neural networks more explainable \cite{samek2021explaining}. This increases trust among healthcare professionals and end users. By simplifying the model's decision-making process, healthcare professionals may use AI-assisted technologies in clinical settings with confidence. The primary objective of our approach is twofold. First, it aids healthcare professionals in expediting the diagnostic process by offering AI-generated insights. Second, it allows individuals to screen suspicious skin lesions themselves. This helps with early detection. If abnormalities are found, users can consult a doctor or dermatologist. This ensures timely treatment and better patient outcomes.
In summary, our key contributions are:
\begin{itemize}
\item  We implement transfer learning and subsequently fine-tune state-of-the-art deep learning models for mpox detection using skin lesion images.
\item We employ XAI techniques to validate model predictions and improve interpretability.
\item We fused the MSLD \cite{ali2022monkeypox} and MSID \cite{bala2023monkeynet} datasets to enhance model generalization.
\item We aim to alleviate the workload of clinicians by automating monkeypox detection from skin lesion images specifically in low-resource settings.
\end{itemize} 

This article is organized into several sections. Section \ref{sec2} provides a summary of related studies. The research approach is explained in Section \ref{sec3}. Section \ref{sec4} describes the experimental setup including dataset details and data preprocessing steps. It also outlines the hyperparameter settings for different models. Section \ref{sec5} presents the research outcomes and compares the performance of various models. Finally, Section \ref{sec6} discusses the study's limitations and suggests directions for future research.

\section{Related Work}\label{sec2}
The studies reviewed present various approaches to detecting and classifying monkeypox with a focus on DL and ML models using different datasets and methodologies. \cite{haque2022human} integrates a CBAM with deep transfer learning. It evaluates models like VGG19, Xception, DenseNet121, EfficientNetB3, MobileNetV2 on the MSLD \cite{ali2022monkeypox} dataset and attains the maximum level of validation accuracy of 83.89\%. It highlights the effectiveness of attention mechanisms in enhancing diagnostic precision although it acknowledges the need for larger datasets and diverse image sources. \cite{syed2022monkey} applies machine learning classifiers (KNN, AdaBoost, Naive Bayes, Decision Tree) on the MSLD \cite{ali2022monkeypox} dataset focusing on early detection. It emphasizes the model's potential but points out the limitations of dataset size and diversity. \cite{gurbuz2022monkeypox} explores DL models including DenseNet121, ResNet-50 \cite{koonce2021resnet}, Xception, EfficientNetB3 and EfficientNetB7 on the MSLD \cite{ali2022monkeypox} dataset with data replication techniques to address dataset limitations. EfficientNetB7 achieved the highest accuracy of 90\%. The study recommends larger and more varied datasets. \cite{sitaula2022monkeypox} compares 13 pre-trained models and custom layers achieving 87.13\% accuracy. It introduces ensemble techniques for improved diagnostic accuracy although it faces challenges with small datasets and memory constraints. \cite{ahsan2022image} introduces the Monkeypox2022 dataset and a modified VGG16 model by achieving 97\% accuracy in one study while another achieved 88\%. It highlights the impact of small datasets on generalizability and calls for larger samples. All studies recognize the limitations of small datasets with suggestions for future work focused on dataset expansion, real-time applications and model robustness.

\cite{ali2022monkeypox} and \cite{ahsan2022transfer} investigate the use of DL algorithms to identify monkeypox lesions but differ in their methodologies, datasets and results. Both studies utilize transfer learning techniques with established architectures including VGG16, ResNet50 \cite{koonce2021resnet} and Inception-based models to address the challenge of limited medical image datasets. \cite{ali2022monkeypox} emphasizes a rather tiny dataset that was assembled using openly accessible sources such as case reports and websites. They achieve the highest accuracy with ResNet50 (82.96\%). In contrast, \cite{ahsan2022transfer} works with a similarly small sample of Monkeypox images but evaluates a broader set of models (VGG16, Inception-ResNetV2, ResNet50 \cite{koonce2021resnet}, ResNet101, MobileNetV2 \cite{sandler2018mobilenetv2} and VGG19). They attained accuracies between 93\% and 99\% with Inception-ResNetV2 and MobileNetV2. Both studies employ data augmentation techniques including flipping and scaling to enhance model performance. \cite{ali2022monkeypox} specifically mentions the use of ensemble models. Limitations identified in both papers include the small size and limited diversity in the datasets. \cite{ali2022monkeypox} noted the absence of patient metadata. The results in both studies highlight the potential of DL for disease detection but they emphasize that in order to increase generalizability and robustness, larger and more varied datasets are required. Future work in both papers suggests expanding datasets and exploring additional DL techniques to enhance diagnostic accuracy and model interpretability. \cite{ahsan2022transfer} integrates LIME \cite{ribeiro2016should} to interpret the predictions, a feature not explored in \cite{ali2022monkeypox}.

\cite{maqsood2023monkeypox} proposes a DL-based approach using modified Inception-ResNet-V2 and NASNet-Large models. They achieve 98.59\% accuracy through feature selection and fusion, although its reliance on public datasets with limited diversity remains a concern. In contrast, \cite{alrusaini2023deep} focuses on multiple CNN models including VGG-16 and achieves 96\% accuracy with PCA for feature selection. It highlights the limitations of a small dataset and suggests enhancing the dataset and exploring alternative AI techniques. \cite{saleh2023human} introduces a two-phase diagnostic strategy with an Improved Binary Chimp Optimization (IBCO) algorithm for feature selection and an ensemble diagnosis model combining various classifiers. This method achieved 98.48\% accuracy but suffers from limited data and a lack of AI in previous diagnostic practices. \cite{thieme2023deep} develops the MPXV-CNN which incorporates a broad dataset including user-generated data and a personalized recommendation system. It yields high accuracy particularly in certain regions but faces dataset biases and the need for demographic diversity. \cite{alhasson2023deep} presents a deep learning-based mobile application using MobileNetV2. It achieves 99\% accuracy and addresses class imbalance with SMOTETomek. While it shows great promise, dataset representativeness and further clinical validation are needed.

\cite{jaradat2023automated} evaluates pretrained models VGG19, VGG16, ResNet50, MobileNetV2 and EfficientNetB3 on a custom dataset. MobileNetV2 achieves the highest accuracy of 98.16\% with excellent performance in precision, recall and F1-score. \cite{pal2023deep} compares CNN, VGG19, ResNet50, Inception v3 and Autoencoder models on a publicly available dataset. Inception v3 reached an accuracy of 96.56\%. \cite{nayak2023detection} focuses on ResNet-18 and SqueezeNet models using a Kaggle dataset. It reports ResNet-18's accuracy of 91.11\% and introduces explainable AI techniques like LIME \cite{ribeiro2016should} to enhance model interpretability. In contrast, \cite{sorayaie2023monkeypox} uses a more diverse dataset for two- and four-class classification. DenseNet201 outperforms others with 97.63\% accuracy in the two-class scenario and incorporates LIME \cite{ribeiro2016should} and Grad-CAM \cite{selvaraju2017grad} for model transparency. All studies share limitations such as small datasets and the need for external validation. Future work across these studies emphasizes expanding datasets, improving model generalizability and exploring more interpretable AI techniques for real-time clinical applications.

The Mpox Classifier \cite{nayak2024mpox} uses a DenseNet-201 based deep transfer learning model and  achieves an impressive accuracy of 99.1\% with the "MSLD V2" dataset. Although this work acknowledges misclassifications, particularly between chickenpox and monkeypox, it also shows the promise of DL in medical imaging. In contrast, the study using smartphone images for Mpox detection \cite{campana2024transfer} leverages Transfer Learning with pre-trained models like VGG16, Xception and DenseNet-169. It achieves promising results despite the challenge of limited dataset diversity. The research emphasizes the potential of mobile-based solutions in low-resource settings. Meanwhile, the web-based detection system \cite{ali2024web} addresses racial diversity by incorporating a skin-color agnostic augmentation technique. It benchmarks multiple models including VGG16 and Xception achieving an accuracy of 83.59\%. The study points out the need for a more diverse and dermatologist-verified dataset. \cite{asif2024ai} synthesizes AI-based methods for Mpox detection focusing on CNNs and ML techniques. It calls for the expansion of datasets and real-world model validation while underscoring AI's potential to increase diagnostic precision. The automated machine learning model \cite{demir2024mnpdensenet} introduces a feature engineering approach with DenseNet201 and SVM achieving 91.87\% accuracy. It faces dataset limitations but offers a significant contribution to medical image analysis with feature selection techniques. All these studies emphasize the need for more diverse, larger datasets and real-world validation to enhance model robustness and accessibility. Future work in all studies involves dataset expansion and model refinement to improve diagnostic accuracy and generalizability.

\cite{yue2024ultrafast} introduces the Fast-MpoxNet model based on ShuffleNetV2. It enhances accuracy and interpretability through attention feature fusion and DropBlock. It is trained on a combination of the MSID \cite{bala2023monkeynet} and the MSLD \cite{ali2022monkeypox} dataset. Images of Mpox and other skin disorders are included in these collections. The model performs well in accuracy and inference speed. However, it faces limitations in dataset diversity and requires future validation with more clinical images. In contrast, \cite{jain2024multi} develops the MPXCN-Net system which combines features from three pre-trained CNNs (DarkNet19, MobileNetV2 and ResNet18) with a SVM for classification. This multi-model approach achieves 90.4\% classification accuracy but still requires validation on diverse datasets to improve generalization. These papers share a common limitation in the need for more diverse datasets and further validation. Future research across these studies could focus on improving data diversity, exploring hybrid models and refining the algorithms for more robust performance in real-world settings.
\\ \\
Our comprehensive literature review highlights the following research gaps:
\begin{itemize}
\item  To increase model generalizability and real-world reliability, more and more varied datasets are required.
\item AI models require better interpretability through explainable techniques for clearer decision-making.
\item Reducing computational constraints is essential for practical deployment especially in low-resource environments.
\item Hybrid and ensemble models along with optimized mobile and web-based solutions can enhance classification accuracy and real-time detection.
\end{itemize}

\section{Methodology}\label{sec3}

Our study began with assembling a comprehensive dataset of skin lesion images. Two publicly accessible datasets, MSLD \cite{ali2022monkeypox} and MSID \cite{bala2023monkeynet}, were utilized. These datasets contained images of confirmed Monkeypox cases along with images of other skin conditions such as measles, chickenpox and regular skin. The lack of images specifically related to monkeypox was one of the main obstacles. To address this, we applied data preprocessing mentioned in Section \ref{subsec2}.

For classification, we have selected models to train from scratch along with pre-trained deep learning models. At first, we employed a vision transformer (ViT) model to train with our dataset from scratch. We then implemented training on the CNN-ViT hybrid and ResNet-ViT hybrid models. After curating the result of these three models trained from scratch we decided to employ pretrained models like ResNet-50, MobileNet-v2 and ViT B16 for feature extraction due to their ability to capture complex patterns in images. These models were fine-tuned by modifying the final layers to match the number of target classes. During training, cross-entropy loss and an adaptive learning rate optimizer were used to improve efficiency and ensure proper convergence.

\begin{figure}[h]
  \centering
  \includegraphics[width=0.95\textwidth]{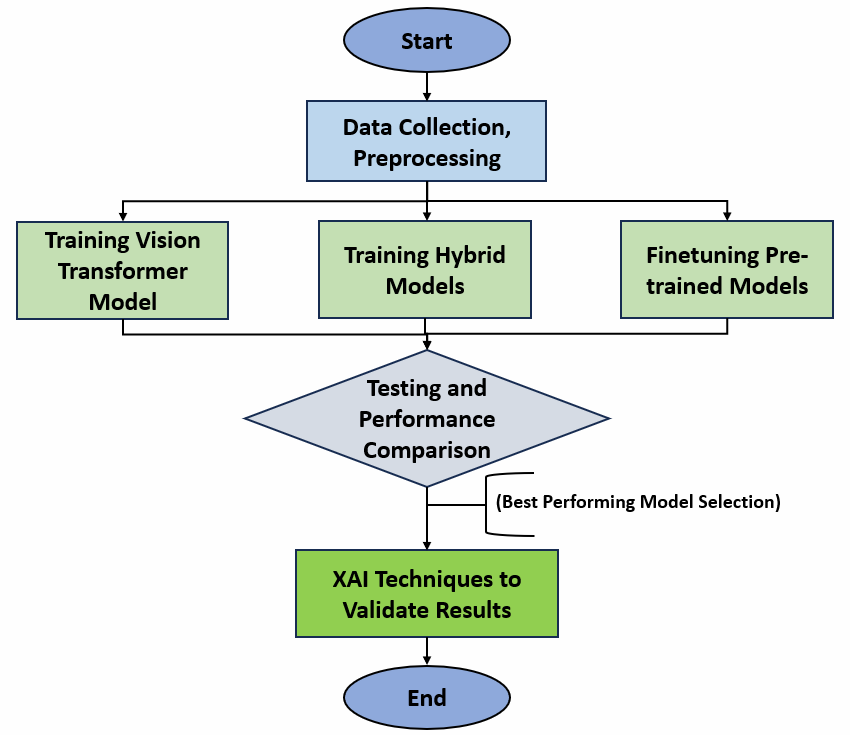}
  \caption{Proposed methodology for Explainable AI-driven Human Monkeypox Detection}\label{fig1}
\end{figure}

We have evaluated the trained models to determine their accuracy in detecting and classifying Monkeypox lesions. To gauge efficacy, performance indicators like F1-score, recall, accuracy and precision were computed. We used a confusion matrix to analyze the classification results and assess how well the models differentiated between Monkeypox and other skin conditions. Fig.\ref{fig1} illustrates the workflow followed in this study.

Finally, explainable AI methods like LIME \cite{ribeiro2016should} and Grad-CAM \cite{selvaraju2017grad} were used to validate the results of the best-performing models. LIME helped in understanding the contribution of individual input features by generating locally interpretable explanations. By emphasizing the significant areas in the images that affected the model's predictions, Grad-CAM offered visual insights. These techniques ensure transparency in decision-making and help assess whether the models focused on relevant areas. By applying these methods, we evaluated the reliability and trustworthiness of the classification results.

\section{Experimental Setup}\label{sec4}

For the experiments we have used an NVIDIA T4 GPU in Google Colab. This GPU provided enough power for training and evaluation. We used standard deep learning libraries in this study. PyTorch was used to build the models. Torchvision helped in fine-tuning pretrained models.

\subsection{Dataset}\label{subsec1}
The dataset for this study came from public image repositories. We fused two main datasets: the MSLD \cite{ali2022monkeypox} and the MSID \cite{bala2023monkeynet}. These datasets included a variety of skin lesion images for the purpose of detecting monkeypox. 

\begin{figure}[h]
  \centering
  \includegraphics[width=0.7\textwidth]{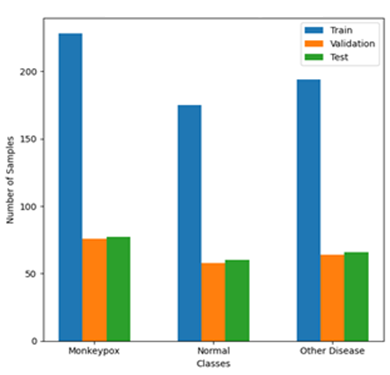}
  \caption{Data distribution into three classes for training, testing and validation}\label{fig2}
\end{figure}

The combined dataset had 998 images each with a 224x224 resolution. The images were divided into three classes: Monkeypox, other skin conditions (Chickenpox, Measles) and normal skin.

\begin{table}[h]
  \caption{Data distribution into three classes for training, testing and validation}\label{Data distribution}
  \begin{tabular}{@{}lccc@{}}
   \toprule
\textbf{Split} & \textbf{Monkeypox} & \textbf{Normal} & \textbf{Other Disease} \\
    \midrule
    Train & 228 & 175 & 194\\
    
    Test & 77 & 60 & 66 \\
    
    Validation & 76 & 58 & 64\\
    \botrule
\end{tabular}
\end{table}

Training, testing and validation partitions of the dataset are predetermined. There are 597 images in the training set, 203 in the test set and 198 in the validation set. The split follows a 60:20:20 ratio. Table. \ref{Data distribution} and Figure \ref{fig2} show the distribution of images among the three classes: Monkeypox, Normal and Other disease.

\subsection{Data Preprocessing}\label{subsec2}
The dataset lacks sample images, especially Monkeypox-specific images. To address this, we meticulously curated both the MSLD \cite{ali2022monkeypox} and MSID \cite{bala2023monkeynet} datasets and carefully fused them to ensure an adequate number of images. Our final dataset included 998 images distributed into three classes (Monkeypox, Normal and Other disease). Due to inconsistencies in image quality across different sources, we resize every image to a set dimension of 224×224. We then systematically separated the images into subsets for testing (20\%), validation (20\%) and training (60\%).

\subsection{Model Hyperparameters}\label{subsec3}

\begin{table}[h]
  \caption{Model Hyperparameters}\label{Model Hyperparameters}
  \begin{tabular}{@{}lcccc@{}}
   \toprule
\textbf{Model} & \multicolumn{1}{c}{\textbf{Number of}} & \textbf{Training Batch} & \textbf{Optimizer} & \textbf{Learning} \\
               & \multicolumn{1}{c}{\textbf{Epochs}}     & \textbf{Size} &                  & \textbf{Rate} \\
    \midrule
    ViT (From Scratch) & 15 & 32 & Adam & 0.0003 \\

    CNN-ViT Hybrid & 11 & 32 & AdamW & 0.0003 \\
    
    ResNet-ViT Hybrid & 20 & 32 & AdamW & 0.00006 \\
    
    ResNet-50 & 16 & 32 & Adam & 0.00008 \\

    MobileNet-v2 & 16 & 32 & Adam & 0.00011 \\

    ViT B16 & 16 & 32 & AdamW & 0.00004 \\
    \botrule
\end{tabular}
\end{table}

Table \ref{Model Hyperparameters} shows the set of hyperparameters used for training different model architectures in our study. ViT model from scratch,  ResNet-50 and  MobileNet-v2 is trained using the Adam optimizer. On the other hand, CNN-ViT Hybrid, ResNet-ViT Hybrid and ViT B16 is trained using the AdamW optimizer. We kept the training batch size fixed at 32 across all models. The number of training epochs varied model to model.  ResNet-50, MobileNet-v2 and  ViT B16 pretrained models are trained for 16 epochs, while ViT (From Scratch) trained for 15 epochs,  CNN-ViT Hybrid trained for 11 epochs and  ResNet-ViT Hybrid trained for 20 epochs. We have also applied hyperparameter tuning strategies for finding the best combination of learning rate for different models and ends up to a learning rate as stated in the Table \ref{Model Hyperparameters}.

\subsection{Explainable AI Techniques}\label{subsec4}
Two well-known explainable AI (XAI) techniques were used to validate and ensure the reliability of the best-performing models. These techniques are Local Interpretable Model-agnostic Explanations (LIME) \cite{ribeiro2016should} and Gradient-weighted Class Activation Mapping (Grad-CAM) \cite{selvaraju2017grad}. LIME explains model predictions by creating locally interpretable approximations of complex models. Grad-CAM highlights important regions in an image by using gradient-based activation maps from convolutional layers.

\section{Result Analysis}\label{sec5} 
We evaluated the performance of three models trained from scratch and three pre-trained models on human monkeypox detection task. The models were compared across multiple metrics including accuracy, weighted average precision, weighted average recall and weighted average F1-score.

\begin{table}[h]
  \caption{Model Evaluation Results}\label{Model Evaluation-1}
  \begin{tabular}{@{}lcccc@{}}
   \toprule
   \textbf{Model} & \multicolumn{1}{c}{} & \textbf{Weighted Average}  &   & \textbf{Accuracy} \\
               & \multicolumn{1}{c}{\textbf{Precision (\%)}}     & \textbf{Recall (\%)} &  \textbf{F1 Score (\%)} &  \textbf{(\%)} \\
    \midrule
    ViT (From Scratch) & 67.56 & 65.02 & 65.28 & 65.02 \\

    CNN-ViT Hybrid & 71.76 & 69.46 & 69.79 & 69.46 \\
    
    ResNet-ViT Hybrid & 66.78 & 67.00 & 65.81 & 67 \\
    
    ResNet-50 & 86.35 & 86.21 & 86.21 & 86.21 \\

    MobileNet-v2 & \textbf{93.33} & \textbf{93.10} & \textbf{93.09} & \textbf{93.15} \\

    ViT B16 & 92.43 & 92.13 & 92.16 & 92.12 \\
    \botrule
\end{tabular}
\end{table}

The results from Table. \ref{Model Evaluation-1} show significant variation in model performance across different architectures. ViT (From Scratch) has the lowest performance with an accuracy of 65.02\%. Its precision, recall, and F1-score are also below 68\% which indicates its limitations in feature extraction and classification. The CNN-ViT Hybrid improves performance compared to the standalone ViT model. It achieves an accuracy of 69.46\% with slightly better precision, recall and F1-score.

The ResNet-ViT Hybrid model does not outperform the CNN-ViT Hybrid. Its accuracy remains at 67\% with marginal improvements in precision and recall. This indicates that combining ResNet with ViT does not provide a significant advantage over other hybrid models. ResNet-50, a well-established CNN model, performs significantly better than the hybrid models. It achieves 86.21\% accuracy and maintains high precision and recall.

MobileNet-v2 achieves the highest performance among all models. It reaches 93.15\% accuracy with consistently high precision, recall and F1-score above 93\%. ViT B16, another transformer-based model, closely follows MobileNet-v2. It attains 92.12\% accuracy with comparable precision, recall and F1-score. This demonstrates that transformer models can achieve competitive performance when properly optimized.\\

\begin{figure}[h]
  \centering
  \includegraphics[width=0.95\textwidth]{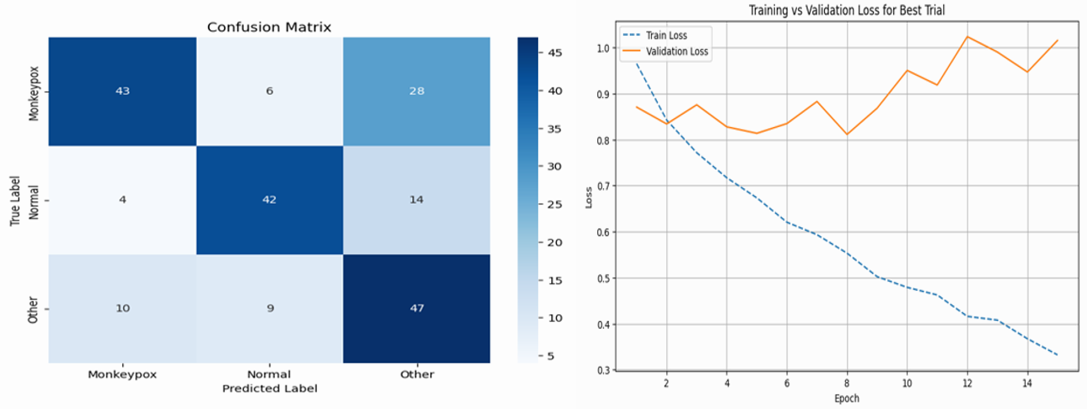}
  \caption{Confusion matrix and Training vs Validation Loss of ViT model trained from scratch}\label{fig3}
\end{figure}

The confusion matrix of Figure \ref{fig3} shows poor classification performance. The model struggles to distinguish between classes, especially for the "Monkeypox" category, which has a high misclassification rate. The "Normal" and "Other disease" classes also have several incorrect predictions which indicate that the model has difficulty learning discriminative features. The training vs. validation loss curve of Figure \ref{fig3} highlights a significant issue. The model is learning from the training data when the training loss steadily drops. However, the validation loss remains high and fluctuates showing no improvement. This implies overfitting, a situation in which the model retains training data but is unable to generalize to new data. The increasing validation loss confirms poor generalization and ineffective learning. This result indicates that the ViT model trained from scratch, struggles to converge and does not perform well for this classification task.

\begin{figure}[h]
  \centering
  \includegraphics[width=0.95\textwidth]{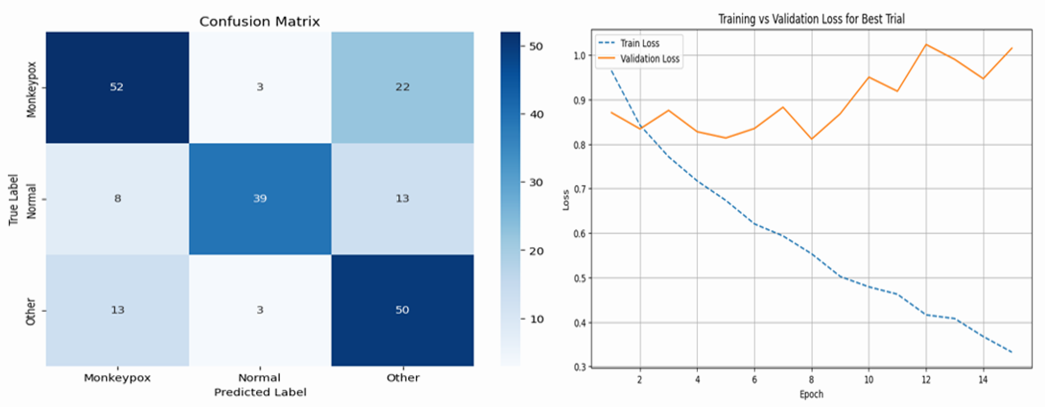}
  \caption{Confusion matrix and Training vs Validation Loss of CNN-ViT Hybrid model trained from scratch}\label{fig4}
\end{figure}

\begin{figure}[h]
  \centering
  \includegraphics[width=0.95\textwidth]{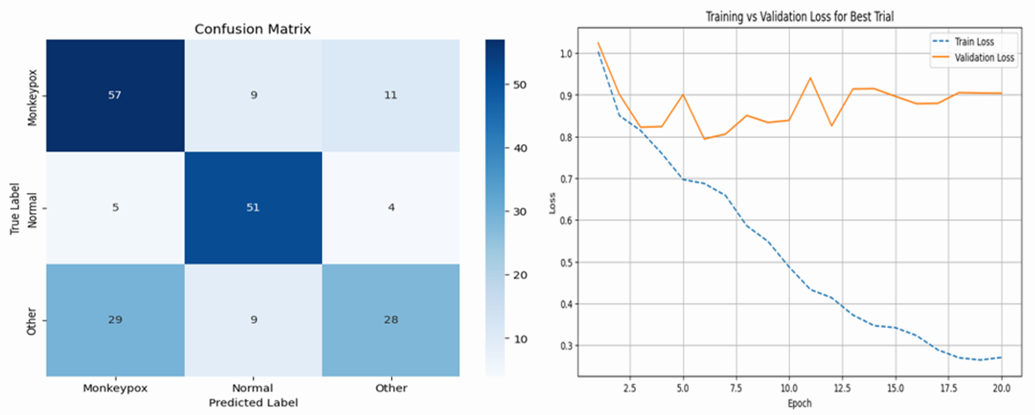}
  \caption{Confusion matrix and Training vs Validation Loss of ResNet-ViT Hybrid model trained from scratch}\label{fig5}
\end{figure}

The CNN-ViT Hybrid model (Figure \ref{fig4}) shows improved classification compared to the standalone ViT model. It correctly classifies more samples across all categories but misclassification remains high, particularly in the "Monkeypox" and "Other disease" classes. The "Normal" class has fewer errors indicating better feature learning. In comparison, the ResNet-ViT Hybrid model (Figure \ref{fig5}) demonstrates moderate classification performance. It performs well in the "Monkeypox" and "Normal" classes but struggles with the "Other disease" class often misclassifying these samples as "Monkeypox." This suggests that distinguishing between these two categories remains a challenge. Both models exhibit overfitting as shown in their training vs. validation loss curves. The CNN-ViT Hybrid model (Figure \ref{fig4}) shows a steady decrease in training loss indicating effective learning from training data. However, validation loss remains high and unstable which suggests poor generalization. Similarly, the ResNet-ViT Hybrid model (Figure \ref{fig5}) also shows steadily decreasing training loss but its validation loss remains high and fluctuates that confirms overfitting. The lack of validation loss improvement indicates difficulty in adapting to unseen data. While both hybrid models provide some benefits over the standalone ViT model, training from scratch still results in suboptimal generalization.

\begin{figure}[h]
  \centering
  \includegraphics[width=0.95\textwidth]{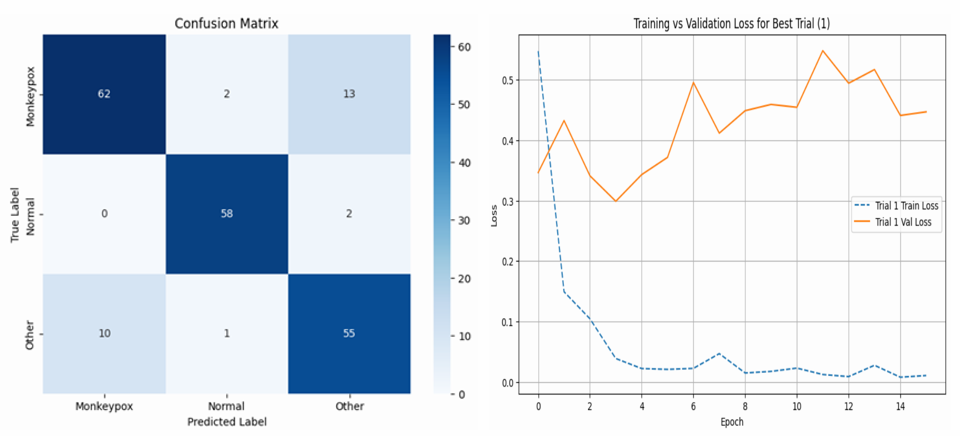}
  \caption{Confusion matrix and Training vs Validation Loss of ResNet-50 pre-trained model}\label{fig6}
\end{figure}

The confusion matrix of Figure \ref{fig6} indicates strong classification performance. The model correctly predicts most samples in all three classes. The "Normal" class has the highest accuracy with very few misclassifications. The "Monkeypox" and "Other disease" classes also show high accuracy with only a small number of errors. The training vs. validation loss curve of Figure \ref{fig6} shows a clear learning trend. The training loss decreases rapidly and stabilizes at a low value. This demonstrates how well the model absorbs the training data. Good generalization is indicated by the validation loss, which stays constant with very slight variations. The small gap between training and validation loss suggests low overfitting. These results confirm that the ResNet-50 model performs well on this classification task.

\begin{figure}[h]
  \centering
  \includegraphics[width=0.95\textwidth]{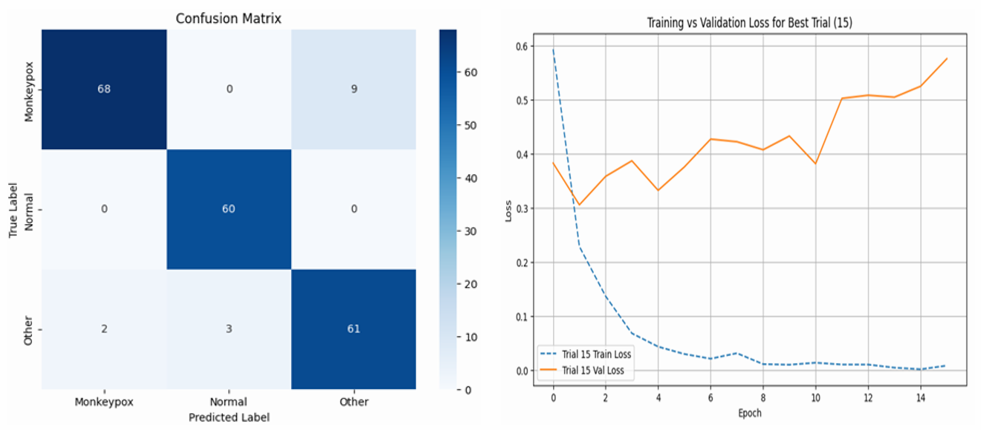}
  \caption{Confusion matrix and Training vs Validation Loss of MobileNet-v2 pre-trained model}\label{fig7}
\end{figure}

\begin{figure}[h]
  \centering
  \includegraphics[width=0.95\textwidth]{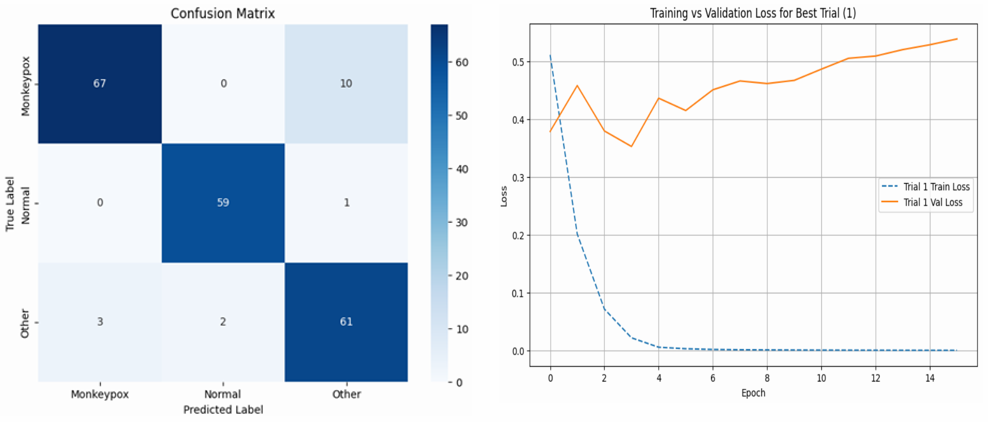}
  \caption{Confusion matrix and Training vs Validation Loss of ViT B16 pre-trained model}\label{fig8}
\end{figure}

The confusion matrices of Figures \ref{fig7} and \ref{fig8} indicate strong classification performance. Both models correctly classify most samples across all three categories. The "Normal" class shows near-perfect accuracy in both cases with Figure \ref{fig7} having no misclassifications and Figure \ref{fig8} having only one. The "Monkeypox" and "Other disease" classes also exhibit high accuracy though some misclassification errors are present. The training vs. validation loss curves for both models show similar trends. The training loss decreases rapidly and stabilizes at a very low value that confirms that both models learn efficiently from the training data. However, the validation loss remains high and fluctuates which suggests overfitting. In Figure \ref{fig8}, the validation loss increases over epochs further indicating poor generalization. Despite this limitation, both models achieve high classification accuracy that demonstrates effective learning but limited adaptability to unseen data.

Overall, transformer-based models alone do not perform well when trained from scratch. Hybrid models offer moderate improvements but do not surpass traditional CNN architectures. ResNet-50 provides strong performance but MobileNet-v2 outperforms all models. ViT B16 performs slightly below MobileNet-v2 and suggests that well-trained transformers can compete with CNN-based models.

With an accuracy of 93.15\%, the comparative analysis identifies MobileNet-v2 as the top-performing model. It consistently achieved superior results across all other evaluation metrics. ViT B16 and ResNet-50 follow the path with promising accuracy of 92.12\% and 86.21\% respectively. Our experiment shows that training models from scratch with such limited dataset is not feasible. Transfer learning is suitable in such cases.

To further validate and bring trustworthiness of our evaluation we employed explainable AI (XAI) techniques like LIME \cite{ribeiro2016should} and Grad-CAM \cite{selvaraju2017grad}. We have selected three best performing models (MobileNet-v2, ViT B16 and ResNet-50) from the model performance evaluation phase. We then applied LIME and Grad-CAM to uncover the underlying reasons behind the model's predictions.

\begin{figure}[h]
  \centering
  \includegraphics[width=0.95\textwidth]{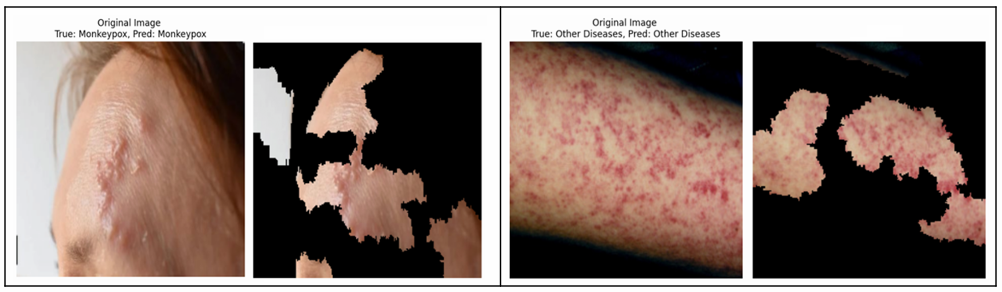}
  \caption{Visualization of LIME applied in MobileNet-v2 models prediction}\label{fig9}
\end{figure}

Figure \ref{fig9} presents a visualization of LIME applied to the predictions of a MobileNet-v2 model for monkeypox detection. The left section of Figure \ref{fig9} shows an image correctly classified as monkeypox with highlighted regions indicating important features contributing to the decision. The right section of Figure \ref{fig9} depicts an image correctly classified as another disease with relevant regions identified by LIME. The features that improved interpretability and confidence in the categorization and impacted the model's choice are indicated by the highlighted areas.

\begin{figure}[h]
  \centering
  \includegraphics[width=0.95\textwidth]{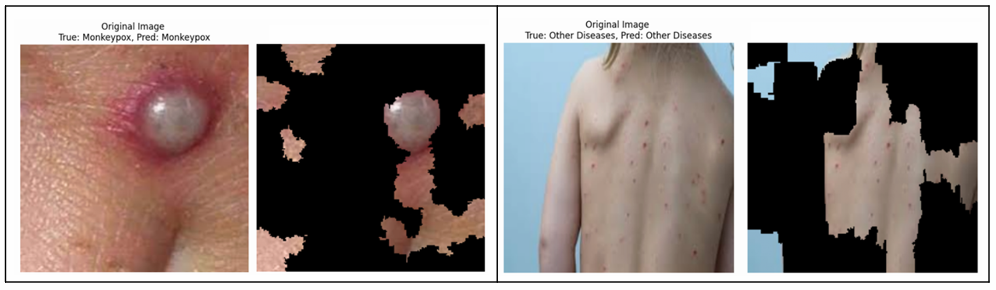}
  \caption{Visualization of LIME applied in ResNet-50 models prediction}\label{fig10}
\end{figure}

Figure \ref{fig10} illustrates the application of LIME on ResNet-50 models predictions for monkeypox detection. The left section of Figure \ref{fig10} shows an image correctly classified as monkeypox with highlighted regions indicating the most influential features. The right section of Figure \ref{fig10} presents an image correctly classified as another disease with key areas contributing to the decision. The highlighted regions help explain the model’s reasoning and improve interpretability and reliability in diagnosis.

\begin{figure}[h]
  \centering
  \includegraphics[width=0.95\textwidth]{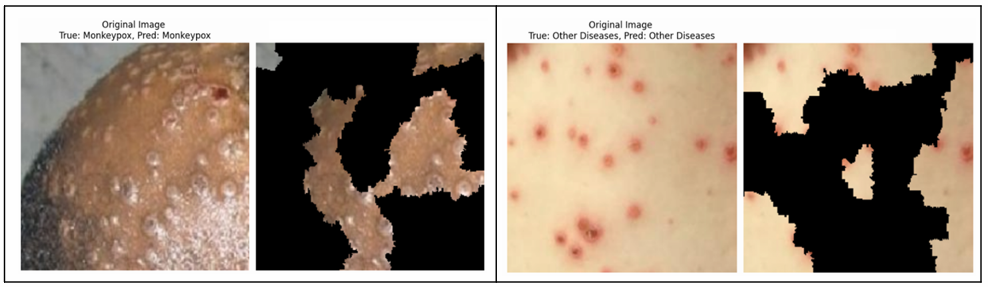}
  \caption{Visualization of LIME applied in ViT B16 models prediction}\label{fig11}
\end{figure}

Figure \ref{fig11} demonstrates the application of LIME on Vision Transformer (ViT-B16) models predictions for monkeypox detection. The left section of Figure \ref{fig11} shows an image correctly classified as monkeypox with highlighted regions indicating important features influencing the decision. The right section of Figure \ref{fig11} presents an image correctly classified as another disease with important factors influencing the model's classification. By offering insights into the ViT-B16 model's process of decision-making, the highlighted regions improve interpretability.

\begin{figure}[h]
  \centering
  \includegraphics[width=0.95\textwidth]{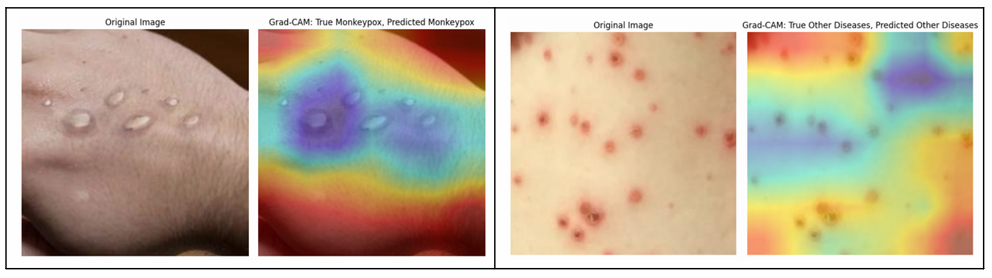}
  \caption{Visualization of Grad-CAM applied in MobileNet-v2 models prediction}\label{fig12}
\end{figure}

Figure \ref{fig12} presents a visualization of Grad-CAM applied to MobileNet-v2 model predictions for monkeypox detection. The left set of Figure \ref{fig12} shows an original image of a monkeypox lesion and its corresponding Grad-CAM heatmap. The heatmap highlights the model's focus areas and confirms a correct prediction. The right set of Figure \ref{fig12} displays an image of a different skin disease and its associated Grad-CAM heatmap. The model correctly identifies it as a non-monkeypox disease.

\begin{figure}[h]
  \centering
  \includegraphics[width=0.95\textwidth]{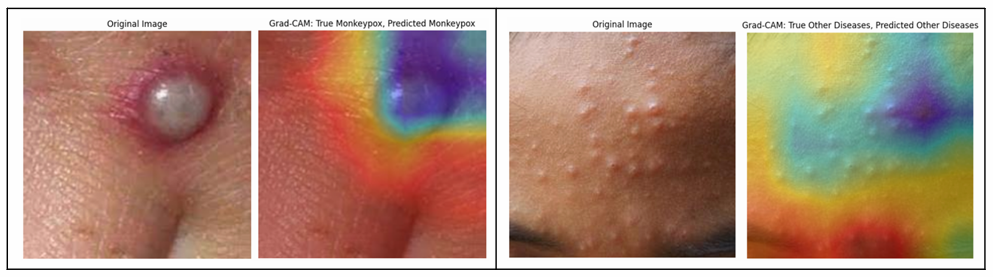}
  \caption{Visualization of Grad-CAM applied in ResNet-50 models prediction}\label{fig13}
\end{figure}

\begin{figure}[h]
  \centering
  \includegraphics[width=0.95\textwidth]{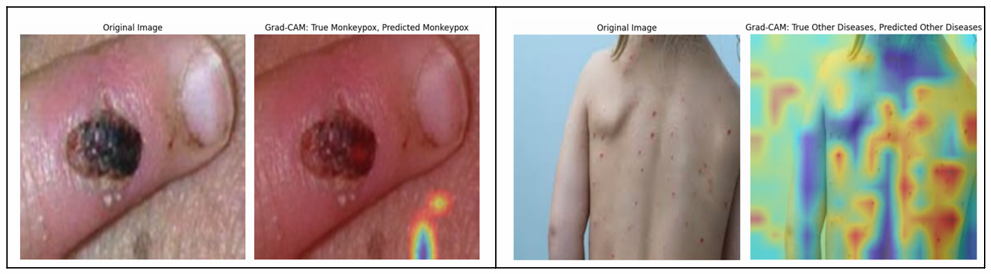}
  \caption{Visualization of Grad-CAM applied in ViT B16 models prediction}\label{fig14}
\end{figure}

Figure \ref{fig13} shows a Grad-CAM visualization applied to ResNet-50 model predictions for monkeypox detection. The left set of Figure \ref{fig13} contains an original image of a monkeypox lesion and its corresponding Grad-CAM heatmap. The highlighted regions indicate the model's focus and  confirm a correct prediction. The right set of Figure \ref{fig13} presents an image of a different skin disease and its Grad-CAM heatmap. The model correctly identifies it as a non-monkeypox disease. This visualization demonstrates how the model identifies key areas for classification and improves the interpretability of deep learning-based medical diagnosis.

Figure \ref{fig14} presents a Grad-CAM visualization applied to ViT-B16 model predictions for monkeypox detection. The left set of Figure \ref{fig14} displays an original image of a monkeypox lesion and its corresponding Grad-CAM heatmap. The highlighted regions indicate the model’s focus and confirm a correct prediction. The right set of Figure \ref{fig14} shows an image of a different skin disease and its Grad-CAM heatmap. The model correctly classifies it as a non-monkeypox disease.

\section{Conclusion and Future Work}\label{sec6} 
Three deep learning and vision transformer-based models that were trained from scratch for the classification of monkeypox disease from skin lesions using the MSLD and MSID datasets are compared in this paper along with an initial feasibility analysis. As there were not enough images in the dataset, the models performed poorly. To get better predictions with this small dataset we employed three pre-trained models and fine-tuned with our dataset. MobileNet-v2 outperformed other models with an accuracy of 93.15\%. Our study shows the effectiveness of transfer learning in case of dataset scarcity. We used LIME and Grad-CAM to find the fundamental causes of the model's predictions in order to further enhance the interpretability of the top-performing models. One of the major drawbacks of this study is the lack of dataset which prohibited us from getting better performance from models trained from scratch. In future we hope to apply dataset augmentation to improve the performance of models trained from scratch and apply federated learning techniques to preserve privacy of patient data.

\backmatter

\bibliography{sn-bibliography}

\end{document}